\documentclass{article}
\usepackage{ijcai23}

\usepackage{times}
\usepackage{soul}
\usepackage{url}
\usepackage[hidelinks]{hyperref}
\usepackage[utf8]{inputenc}
\usepackage[small]{caption}
\usepackage{graphicx}
\usepackage{amsmath}
\usepackage{amsthm}
\usepackage{booktabs}
\usepackage{algorithm}
\usepackage{algorithmic}
\usepackage[switch]{lineno}
\usepackage{multirow}
\usepackage{amssymb}
\usepackage{diagbox}
\title{Co-training with High-Confidence Pseudo Labels for Semi-supervised Medical Image Segmentation}

\author{
Zhiqiang Shen$^{1,2}$
\and
Peng Cao$^{1,2}$\thanks{Corresponding author}\and
Hua Yang$^{3}$ \and
Xiaoli Liu $^{4}$ \and
Jinzhu Yang$^{1,2}$ \and
Osmar R. Zaiane$^{5}$
\affiliations
$^1$College of Computer Science and Engineering, Northeastern University, Shenyang, China\\
$^2$Key Laboratory of Intelligent Computing in Medical Image, Ministry of Education, Shenyang, China\\
$^3$College of Photonic and Electronic Engineering, Fujian Normal University, Fuzhou, China \\
$^4$DAMO Academy, Alibaba Group, China \\
$^5$Alberta Machine Intelligence Institute, University of Alberta, Edmonton, Alberta, Canada
\emails
xxszqyy@gmail.com,
caopengneu@gmail.com
}

\begin{document}

\maketitle

\begin{abstract}
    Consistency regularization and pseudo labeling-based semi-supervised methods perform co-training using the pseudo labels from multi-view inputs. However, such co-training models tend to converge early to a consensus, degenerating to the self-training ones, and produce low-confidence pseudo labels from the perturbed inputs during training. 
    To address these issues, we propose an \textbf{U}ncertainty-guided \textbf{C}ollaborative \textbf{M}ean-\textbf{T}eacher (UCMT) for semi-supervised semantic segmentation with the high-confidence pseudo labels. Concretely, UCMT consists of two main components: 1) collaborative mean-teacher (CMT) for encouraging model disagreement and performing co-training between the sub-networks, and 2) uncertainty-guided region mix (UMIX) for manipulating the input images according to the uncertainty maps of CMT and facilitating CMT to produce high-confidence pseudo labels. 
    Combining the strengths of UMIX with CMT, UCMT can retain model disagreement and enhance the quality of pseudo labels for the co-training segmentation.
    Extensive experiments on four public medical image datasets including 2D and 3D modalities demonstrate the superiority of UCMT over the state-of-the-art. 
    Code is available at: \href{https://github.com/Senyh/UCMT}{https://github.com/Senyh/UCMT}.
\end{abstract}

\section{Introduction}
Semantic segmentation is critical for medical image analysis. Great progress has been made by deep learning-based segmentation models relying on a large amount of labeled data \cite{chen2018encoder,ronneberger2015u}. However, labeling such pixel-level annotations is laborious and requires expert knowledge especially in medical images, resulting in that labeled data are expensive or simply unavailable. Unlabeled data, on the contrary, are cheap  and relatively easy to obtain. Under this condition, semi-supervised learning (SSL) has been the dominant data-efficient strategy through exploiting information from a limited amount labeled data and an arbitrary amount of unlabeled data, so as to alleviate the label scarcity problem \cite{van2020survey}.

\begin{figure}[!t]
	\centering
	\includegraphics[width=3.4 in]{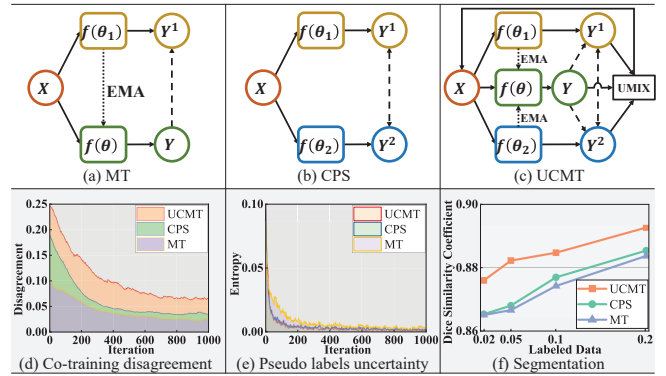}
	\caption{Illustration of the architectures and curves for co-training based semi-supervised semantic segmentation. (a) Mean-teacher \protect\cite{tarvainen2017mean}, (b) Cross pseudo supervision \protect\cite{chen2021semi}, (c) Uncertainty-guided collaborative mean-teacher, 
  (d) the disagreement variation between the co-training sub-networks in terms of dice loss (\textit{w.r.t.} number of iterations) of two branches ($Y^1$ and $Y$ in MT; $Y^1$ and $Y^2$ in CPS; $Y^1$ and $Y^2$ in UCMT), 
 (e) the uncertainty variation of the pseudo labels in terms of entropy \textit{w.r.t.} number of iterations, and (f) the performance of MT, CPS, and UCMT on semi-supervised skin lesion segmentation under different ratio of labeled data. EMA: exponential moving average.}
\label{FIG:CPS2UCMT}
\end{figure}

Consistency regularization \cite{tarvainen2017mean} and pseudo labeling \cite{lee2013pseudo} are the two main methods for semi-supervised   semantic segmentation. Currently, combining consistency regularization and pseudo labeling via cross supervision between the sub-networks, has shown promising performance for semi-supervised segmentation \cite{lee2013pseudo,ouali2020semi,ke2020guided,chen2021semi,liu2022perturbed}. One critical limitation of these approaches is that the sub-networks tend to converge early to a consensus situation causing the co-training model degenerating to the self-training \cite{yu2019does}.
Disagreement between the sub-networks is crucial for co-training, where the sub-networks initialized with different parameters or trained with different views have different biases (i.e., disagreement) ensuring that the information they provide is complementary to each other.
Another key factor affecting the performance of these approaches is the quality of pseudo labels. 
Intuitively, high quality pseudo labels should have low uncertainty \cite{grandvalet2004semi}.  More importantly, these two factors influence each other. Increasing the degree of the disagreement between the co-training sub-networks by different perturbations or augmentations could result in their opposite training directions, thus increasing the uncertainty of pseudo labels. To investigate the effect of the disagreement and the quality of pseudo labels for co-training based semi-supervised segmentation, which has not been studied in the literature, we conduct a pilot experiment to illustrate these correlations. As shown in Figure \ref{FIG:CPS2UCMT}, compared with mean-teacher (MT) \cite{tarvainen2017mean} [Figure \ref{FIG:CPS2UCMT} (a)], cross pseudo supervision (CPS) \cite{chen2021semi} [Figure \ref{FIG:CPS2UCMT} (b)] with the higher model disagreement [(d)] and the lower uncertainty [Figure \ref{FIG:CPS2UCMT} (e)] produces higher performance [Figure \ref{FIG:CPS2UCMT} (f)] on semi-supervised segmentation. Note that the dice loss of two branches are calculated to measure the disagreement. The question that comes to mind is: \textit{how to effectively improve the disagreement between the co-training sub-networks and the quality of pseudo labels jointly in a unified network for SSL}.

 In this paper, we focus on two major goals: maintaining  model disagreement and the high-confidence pseudo labels at the same time.
To this end, we propose the Uncertainty-guided Collaborative Mean Teacher (UCMT) framework that is capable of retaining higher disagreement between the co-training segmentation sub-networks [Figure \ref{FIG:CPS2UCMT} (d)] based on the higher confidence pseudo labels [Figure \ref{FIG:CPS2UCMT} (e)], thus achieving better semi-supervised segmentation performance under the same backbone network and task settings [Figure \ref{FIG:CPS2UCMT} (f)].
Specifically, UCMT involves two major components: 1) collaborative mean-teacher (CMT), and 2) uncertainty-guided region mix (UMIX), where UMIX operates the input images according to the uncertainty maps of CMT while CMT performs co-training under the supervision of the pseudo labels derived from the UMIX images. 
Inspired by the co-teaching \cite{han2018co,yu2019does,chen2021semi} for struggling with early converging to a consensus situation and degrading into self-training, we introduce a third component, the teacher model, into the co-training framework as a regularizer to construct CMT for more effective SSL. The teacher model acts as self-ensemble by averaging the student models, serving as a third part to guide the training of the two student models. 
Further, we develop UMIX to construct high-confident pseudo labels and perform regional dropout for learning robust semi-supervised semantic segmentation models.
Instead of random region erasing or swapping \cite{devries2017improved,yun2019cutmix}, UMIX manipulates the original image and its corresponding pseudo labels according to the epistemic uncertainty of the segmentation models, which not only reduces the uncertainty of the pseudo labels but also enlarges the training data distribution. 
Finally, by combining the strengths of UMIX with CMT, the proposed approach UCMT significantly improves the state-of-the-art (sota) results in semi-supervised segmentation on multiple benchmark datasets. For example, UCMT and UCMT(U-Net) achieve 88.22\% and 82.14\% Dice Similarity Coefficient (DSC) on ISIC dataset under 5\% labeled data, outperforming our baseline model CPS \cite{chen2021semi} and the state-of-the-art UGCL \cite{wang2022uncertainty} by 1.41\% and 9.47\%, respectively.

In a nutshell, our contributions mainly include:
\begin{itemize}
\item We pinpoint the problem in existing co-training based semi-supervised segmentation methods: the insufficient disagreement among the sub-networks and the lower-confidence pseudo labels. To address the problem, we design an uncertainty-guided collaborative mean-teacher to maintain co-training with high-confidence pseudo labels, where we incorporate CMT and UMIX into a holistic framework for semi-supervised medical image segmentation.
\item To avoid introducing noise into the new samples, we propose an uncertainty-guided regional mix algorithm, UMIX, encouraging the segmentation model to yield high-confident pseudo labels and enlarge the training data distribution.
\item We conduct extensive experiments on four public medical image segmentation datasets including 2D and 3D scenarios. Comprehensive results demonstrate the effectiveness of each component of our method and the advantage of UCMT over the state-of-the-art.
\end{itemize}

\section{Related Work}

\subsection{Semi-Supervised Learning}
Semi-supervised learning aims to improve performance in supervised learning by utilizing information generally associated with unsupervised learning, and vice versa \cite{van2020survey}. A common form of SSL is introducing a regularization term into the objective function of supervised learning to leverage unlabeled data. From this perspective, SSL-based methods can be divided into two main lines, i.e., pseudo labeling and consistency regularization.
Pseudo labeling attempts to generate pseudo labels similar to the ground truth, for which models are trained as in supervised learning \cite{lee2013pseudo}.
Consistency regularization enforces the model's outputs to be consistent for the inputs under different perturbations \cite{tarvainen2017mean}.
Current state-of-the-art approaches have incorporated these two strategies and shown convincing performance for semi-supervised image classification \cite{sohn2020fixmatch,zhang2021flexmatch}. 
Based on this line of research, we explore more effective consistency learning algorithms for semi-supervised segmentation.

\subsection{Semi-Supervised Semantic Segmentation}
Compared  with image  classification, semantic  segmentation requires much more intensively and costly labeling for pixel-level annotations. Semi-supervised semantic segmentation inherits the main ideas of semi-supervised image classification. The combination of consistency regularization and pseudo labeling, mainly conducting cross supervision between sub-networks using pseudo labels, has become the mainstream strategy for semi-supervised semantic segmentation in both natural images \cite{ouali2020semi,chen2021semi} and medical images \cite{yu2019uncertainty,luo2021semi,wu2021semi,wu2022mutual}. 
Specifically, these combined approaches enforce the consistency of the predictions under different perturbations, such as input perturbations \cite{li2020transformation,tu2022guidedmix}, feature perturbations \cite{ouali2020semi}, and network perturbations \cite{tarvainen2017mean,chen2021semi,wu2021semi,wu2022mutual}.
In addition, adversarial learning-based methods, rendering the distribution of model predictions from labeled data to be aligned with those from unlabeled data, can also be regarded as a special form of consistency regularization \cite{hung2018adversarial,li2020shape}. 
However, such cross supervision models may converge early to a consensus, thus degenerating to self-training ones. We hypothesize that enlarging the disagreement for the co-training models based on the high-confidence pseudo labels can improve the performance of SSL.
Therefore, we propose a novel SSL framework, i.e., UCMT, to generate more accurate pseudo labels and maintain co-training for semi-supervised medical image segmentation.

\subsection{Uncertainty-Guided Semi-Supervised Semantic Segmentation} 
Model uncertainty (epistemic uncertainty) can guide the SSL models to capture information from the pseudo labels.
Two critical problems for leveraging model uncertainty are how to \textit{obtain} and \textit{exploit} model uncertainty. Recently, there are mainly two strategies to estimate model uncertainty: 1) using Monte Carlo dropout \cite{gal2016dropout}, and 2) calculating the variance among different predictions \cite{zheng2021rectifying}. For semi-supervised semantic segmentation, previous works exploit model uncertainty to re-weight the training loss \cite{yu2019uncertainty} or selecting the contrastive samples \cite{wang2022uncertainty}. However, these methods require manually setting a threshold to neglect the low-confidence pseudo labels, where the fixed threshold is hard to determine.
In this paper, we \textit{obtain} the epistemic uncertainty by the entropy of the predictions of CMT for the same input and \textit{exploit} the uncertainty to guide the region mix for gradually exploring information from the unlabeled data.

\begin{figure}
	\centering
	\includegraphics[width=3.4 in]{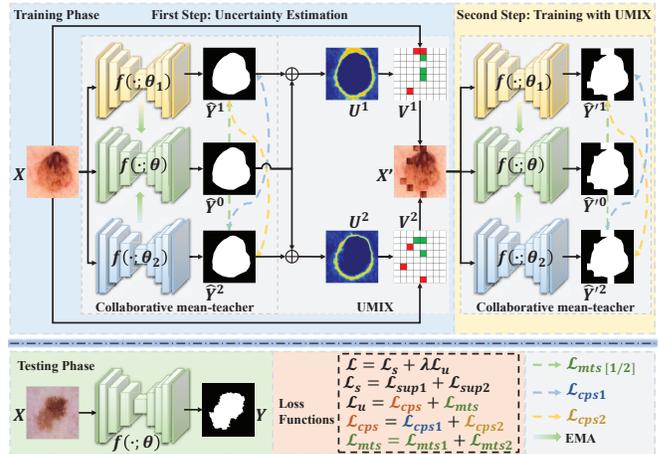}
	\caption{Overview of the proposed UCMT. CMT includes three sub-networks, i.e., the teacher sub-network ($f(\cdot;\theta)$) and the two student sub-networks ($f(\cdot;\theta_1)$ and $f(\cdot;\theta_2)$). UMIX constructs each new samples $X’$ by replacing the top $k$ most uncertain regions (red grids in $V^1$ and $V^2$) with the top $k$ most certain regions (green grids in $V^2$ and $V^1$) in the original image $X$. 
 }
\label{FIG:UMIX+CMT}
\end{figure}

\begin{algorithm}[tb]
    \caption{UCMT algorithm}
    \label{alg:umix+cmt}
    \textbf{Input}: {$\mathcal{D_L} = \{\{(X_i, Y_i)\}_{i=1}^{N}\}, \mathcal{D_U} = \{\{X_j\}_{j=N+1}^{M}\}$}\\
    \textbf{Parameter}: $\theta$, $\theta_1$, $\theta_2$\\
    \textbf{Output}: $f(\cdot; \theta)$
    \begin{algorithmic}[1] 
        \FOR{$T \in [1, numepochs]$}
            \FOR{each minibatch $B$}
                \STATE {//  $i$/$j$ is the index for labeled/unlabeled data}
                \STATE {\textbf{step 1: uncertainty estimation}}
                \STATE {$\hat{Y}^0_i \gets f(X_i; \theta)$,$\hat{Y}^1_i \gets f(X_i; \theta_1)$,$\hat{Y}^2_i \gets f(X_i; \theta_2)$}
                \STATE {$\hat{Y}^0_j \gets f(X_j; \theta)$,$\hat{Y}^1_j \gets f(X_j; \theta_1)$,$\hat{Y}^2_j \gets f(X_j; \theta_2)$}
                \STATE {$L \gets \mathcal{L}_{s}(\hat{Y}^1_i, \hat{Y}^2_i, Y_i) + \lambda(T)\mathcal{L}_{u}(\hat{Y}^0_j, \hat{Y}^1_j, \hat{Y}^2_j)$}
                \STATE {Update $f(\cdot; \theta)$, $f(\cdot; \theta_1)$, $f(\cdot; \theta_2)$ using optimizer}
                \STATE {$U^{[1/2]}_i \gets Uncertain(f(X_i; \theta_{[1/2]}]), f(X_i; \theta))$} 
                \STATE {$U^{[1/2]}_j \gets Uncertain(f(X_j; \theta_{[1/2]}), f(X_j; \theta))$} 
                \STATE {\textbf{step 2: training with UMIX}}
                \STATE {$X'_i / Y'_i \gets$ UMIX$(X_i / Y_i, U^1_i, U^2_i; k, 1/r)$}
                \STATE {$X'_j / \hat{Y'}^0_j  \gets$ UMIX$(X_j / \hat{Y}^0_j, U^1_j, U^2_j; k, 1/r)$}
                \STATE {\textbf{Repeat} 5-8 using $X'_i$, $X'_j$, $ Y'_i$, and $\hat{Y'}^0_j$}
            \ENDFOR
        \ENDFOR
        \STATE \textbf{return} $f(\cdot; \theta)$
    \end{algorithmic}
\end{algorithm}

\section{Methodology}
Before introducing our method, we first define the notations used in this work. The training set $\mathcal{D} = \{\mathcal{D_L}, \mathcal{D_U}\}$ contains a labeled set $\mathcal{D_L} = \{(X_i, Y_i)_{i=1}^{N}\}$ and a unlabeled set $\mathcal{D_U} = \{(X_j)_{j=N+1}^{M}\}$, where $X_i$/$X_j$ denotes the $i_{th}$/$j_{th}$ labeled/unlabeled image, $Y_i$ is the ground truth of the labeled image, and $N$ and $M-N$ are the number of labeled and unlabeled samples, respectively. Given the training data $\mathcal{D}$, the goal of semi-supervised semantic segmentation is to learn a model $f(\cdot; \theta)$ performing well on unseen test sets.


\subsection{Overview}
To avoid the co-training degrading to the self-training, we propose to encourage model disagreement during training and ensure pseudo labels with low uncertainty.
With this motivation, we propose uncertainty-guided collaborative mean-teacher for semi-supervised image segmentation, which includes 1) collaborative mean-teacher, and 2) uncertainty-guided region mix.
As shown in Figure \ref{FIG:CPS2UCMT} (d), CMT and UCMT gradually enlarge the disagreement between the co-training sub-networks. Meanwhile, CMT equipped with UMIX guarantees low-uncertainty for the pseudo labels.
With the help of these conditions, we can safely maintain the co-training status to improve the effectiveness of SSL for exploring unlabeled data.
Figure \ref{FIG:UMIX+CMT} illustrates the schematic diagram of the proposed UCMT. Generally, there are two steps in the training phase of UCMT. In the first step, we train CMT using the original labeled and unlabeled data to obtain the uncertainty maps; Then, we perform UMIX to generate the new samples based on the uncertainty maps. In the second step, we re-train CMT using the UMIX samples. 
Details of the training process of UCMT are shown in Algorithm \ref{alg:umix+cmt}.
Although UCMT includes three models, i.e., one teacher model and two student models, only the teacher model is required in the testing stage.

\subsection{Collaborative Mean-Teacher}
\label{subsec:cmt}
Current consistency learning-based SSL algorithms, e.g., Mean-teacher \cite{tarvainen2017mean} and CPS \cite{chen2021semi}, suggest to perform consistency regularization among the pseudo labels in a multi-model architecture rather than in a single model. However, during the training process, the two-network SSL framework may converge early to a consensus and the co-training degenerate to the self-training \cite{yu2019does}. To tackle this issue, we design the collaborative mean teacher (CMT) framework  by introducing a "arbitrator", i.e., the teacher model, into the co-training architecture \cite{chen2021semi} to guide the training of the two student models. As shown in Figure \ref{FIG:UMIX+CMT}, CMT consists of one teacher model $f(\cdot; \theta)$  and two student models $f(\cdot; \theta_1)$ and $f(\cdot; \theta_2)$, where the teacher model is the self-ensemble of the average of the student models. These models have the same architecture but initialized with different weights for network perturbations. For labeled data, these models are all optimized by supervised learning. For unlabeled data, there are two critical factors: 1) co-training between the two student models, and 2) direct supervision from the teacher to the student models. 

To explore both the labeled and unlabeled data, the total loss $\mathcal{L}$ for training UCMT involves two parts, i.e., the supervised loss $\mathcal{L}_s$ and the unsupervised loss $\mathcal{L}_u$.
\begin{equation}
\mathcal{L} = \mathcal{L}_{s} + \lambda  \mathcal{L}_{u},
\label{loss:total}
\end{equation}
where $\lambda$ is a regularization parameter to balance the supervised and unsupervised learning losses. We adopt a Gaussian ramp-up function to gradually increase the coefficient, i.e., $\lambda(t) = \lambda_m\times\exp{[-5(1-\frac{t}{t_m})^2]}$, where $\lambda_m$ scales the maximum value of the weighted  function, $t$ denotes the current iteration, and $t_m$ is the maximum iteration in training.

\textbf{Supervised Learning Path.} For the labeled data, the supervised loss is formulated as, 
\begin{equation}
\mathcal{L}_{s} = \frac{1}{N} \sum_{i=1}^{N} \big\{ \mathcal{L}_{seg} \left(f\left(X_i;\theta_1\right), Y_i\right) + \mathcal{L}_{seg}\left(f\left(X_i;\theta_2\right), Y_i\right)\big\},
\label{loss:s}
\end{equation}
where $\mathcal{L}_{seg}$ can be any supervised semantic segmentation loss, such as cross entropy loss and dice loss. Note that we choose dice loss in our experiments as its compelling performance in medical image segmentation. 

\textbf{Unsupervised Learning Path.} The unsupervised loss $\mathcal{L}_u$ acts as a regularization term to explore potential knowledge for the labeled and unlabeled data. $\mathcal{L}_u$ includes the \textbf{c}ross \textbf{p}seudo \textbf{s}upervision $\mathcal{L}_{cps}$ between the two student models and the \textbf{m}ean-\textbf{t}eacher \textbf{s}upervision $\mathcal{L}_{mts}$ for guiding the student models from the teacher, as follow:
\begin{equation}
\mathcal{L}_u = \mathcal{L}_{cps} + \mathcal{L}_{mts}.
\label{loss:u}
\end{equation}

\textbf{1) Cross Pseudo Supervision.} The aim of $\mathcal{L}_{cps}$ is to promote two students to learn from each other and to enforce the consistency between them. Let $\mathcal{L}_{cps} = \mathcal{L}_{cps1} + \mathcal{L}_{cps2}$ for encouraging bidirectional interaction for the two student sub-networks $f(\cdot; \theta_1)$ and $f(\cdot; \theta_2)$. The losses of $\mathcal{L}_{cps1}$ and  $\mathcal{L}_{cps1}$ are defined as:
\begin{equation}
\mathcal{L}_{cps[1/2]}= \frac{1}{M-N} \sum_{j=1}^{M-N} \mathcal{L}_{seg} \left(f\left(X_j;\theta_{[1/2]}\right), \hat{Y}_j^{[2/1]}\right),
\label{loss:cps}
\end{equation}
where $\hat{Y}^1_j$ and $\hat{Y}^2_j$ are the pseudo segmentation maps for $X_j$ predicted by $f\left(\cdot;\theta_1\right)$ and $f\left(\cdot;\theta_1\right)$ , respectively.

\textbf{2) Mean-Teacher Supervision.} To avoid the two students co-training in the wrong direction, we introduce a teacher model to guide the optimization of the student models. Specifically, the teacher model is updated by the exponential moving average (EMA) of the average of the student models:
$\theta^t = \alpha \theta^{t-1} + (1-\alpha) [\beta \theta_1^t + (1-\beta)\theta_2^t]$,
where $t$ represents the current training iteration and $\alpha$ is the EMA decay that controls the parameters' updating rate. 

The loss of mean-teacher supervision $\mathcal{L}_{mts} = \mathcal{L}_{mts1} + \mathcal{L}_{mts2}$ is calculated from two branches:
\begin{equation}
\mathcal{L}_{mts[1/2]}= \frac{1}{M-N} \sum_{j=1}^{M-N} \mathcal{L}_{seg} \left(f\left(X_j;\theta_{[1/2]}\right), \hat{Y}^0_j\right),
\label{loss:cmt}
\end{equation}
where $\hat{Y}^0_j$ is the segmentation map derived from $f\left(X_j;\theta\right)$.

\subsection{Uncertainty-Guided Mix}
\label{subsec:umix}
Although CMT can promote model disagreement for co-training, it also slightly increases the uncertainty of the pseudo labels as depicted in Figure \ref{FIG:CPS2UCMT}. 
On the other hand, random regional dropout can expand the training distribution and improve the generalization capability of models \cite{devries2017improved,yun2019cutmix}. However, such random perturbations to the input images inevitably introduce noise into the new samples, thus deteriorating the quality of pseudo labels for SSL. One sub-network may provide some incorrect pseudo labels to the other sub-networks, degrading their performance. 
To overcome these limitations, we propose UMIX to manipulate image patches under the guidance of the uncertainty maps produced by CMT. The main idea of UMIX is constructing a new sample by replacing the top $k$ most uncertain (low-confidence) regions with the top $k$ most certain (high-confidence) regions in the input image. 
As illustrated in Figure \ref{FIG:UMIX+CMT}, UMIX constructs a new sample $X'$ = UMIX$(X, U^1, U^2; k, 1/r)$ by replacing the top $k$ most uncertain regions (red grids in $V^1$ and $V^2$) with the top $k$ most certain regions (green grids in $V^2$ and $V^1$) in $X$, where each region has size $1/r$ to the image size. 
To ensure the reliability of the uncertainty evaluation, we obtain the uncertain maps by integrating the outputs of the teacher and the student model instead of performing $T$ stochastic forward passes designed by Monte Carlo Dropout estimate model \cite{gal2016dropout,yu2019uncertainty}, which is equivalent to sampling predictions from the previous and current iterations.
This process can be formulated as:
\begin{equation}
\begin{aligned}
& U^m = Uncertain(f(X; \theta_m), f(X; \theta)) =  - \sum_{c} P_c \log(P_c), \\
& P_c = \frac{1}{2}(Softmax(f(X; \theta_m)) + Softmax(f(X; \theta))),
\end{aligned}
\end{equation}
where $m=1,2$ denotes the index of the student models and $c$ refers to the class index.

\section{Experiments and Results}

\subsection{Experiments Settings}
\textbf{Datasets.}
We conduct extensive experiments on different medical image segmentation tasks to evaluate the proposed method, including skin lesion segmentation from dermoscopy images, polyp segmentation from colonoscopy images, and the 3D left atrium segmentation from cardiac MRI images.

\textbf{Dermoscopy.} We validate our method on the ISIC dataset \cite{codella2018skin} including 2594 dermoscopy images and corresponding annotations. Following \cite{wang2022uncertainty}, we adopt 1815 images for training and 779 images for validation. 

\textbf{Colonoscopy.} We evaluate the proposed method on the two public colonoscopy datasets, including Kvasir-SEG \cite{jha2020kvasir} and CVC-ClinicDB \cite{bernal2015wm}. Kvasir-SEG and CVC-ClinicDB contain 1000 and 612 colonoscopy images with corresponding annotations, respectively. 

\textbf{Cardiac MRI.} We evaluate our method on the 3D left atrial (LA) segmentation  challenge dataset, which consists of 100 3D gadolinium-enhanced magnetic resonance images and LA segmentation masks for training and validation. Following \cite{yu2019uncertainty}, we split the 100 scans into 80 samples for training and 20 samples for evaluation.


\subsubsection{Implementation Details}
We use DeepLabv3+ \cite{chen2018encoder} equipped with ResNet50 as the baseline architecture for 2D image segmentation, whereas adopt VNet \cite{milletari2016v} as the baseline in the 3D scenario. All images are resized to $256 \times 256$ for inference, while the outputs are recovered to the original size for evaluation, in the 2D scenario.
For 3D image segmentation, we randomly crop $80 \times 112 \times 112 (Depth \times Height \times Width)$ patches for training and iteratively crop patches using a sliding window strategy to obtain the final segmentation mask for testing. We empirically set $\lambda_m=1$, $k=2$, $r=16$, $\alpha=0.99$ and $\beta=0.99$ for our method in the experiments. We implement our method using PyTorch framework on a NVIDIA Quadro RTX 6000 GPU. We adopt AdamW as an optimizer with the fixed learning rate of le-4. The batchsize is set to 16, including 8 labeled samples and 8 unlabeled samples. All 2D models are trained for 50 epochs, while the 3D models are trained for 1000 epochs \footnote{Since UCMT performs the two-step training within one iteration, it is trained for half of the epochs.}.

\subsection{Comparison with State of the Arts}
We compare the proposed method with state-of-the art on the four public medical image segmentation datasets. We re-implement MT \cite{tarvainen2017mean}, CCT \cite{ouali2020semi}, and CPS \cite{chen2021semi} by adopting implementations from \cite{chen2021semi}. For other approaches, we directly use the results reported in their original papers.

\textbf{Results on Dermoscopy.} In Table \ref{Tab:isic}, we report the results of our methods on ISIC and compare them with other state-of-the-art approaches. UCMT substantially outperforms all previous methods and sets new state-of-the-art of 88.22\% DSC and 88.46 DSC under 5\% and 10\% labeled data. For fair comparison with UGCL \cite{wang2022uncertainty}, replace the backbone of UCMT with U-Net.  The results indicate that our UCMT(U-Net) exceeds UGCL by a large margin. Moreover, our CMT version also outperforms other approaches under the two labeled data rates. For example, CMT surpasses MT and CPS by 1.19\% and 1.08\% on 5\% $\mathcal{D_L}$ labeled data, showing the superiority of collaborative mean-teacher against the current consistency learning framework. By introducing UMIX, UCMT consistently increases the performance under different labeled data rates, which implies that promoting model disagreement and guaranteeing high-confident pseudo labels are beneficial for semi-supervised segmentation. 

\begin{table}[!t]
\begin{tabular}{lllll}
\toprule[1pt]
Method & 5\% $\mathcal{D_L}$ & 10\% $\mathcal{D_L}$ \\ \midrule
MT \cite{tarvainen2017mean}     &    86.67       &     87.42       \\
CCT \cite{ouali2020semi}     &     83.97      &      86.43       \\
CPS \cite{chen2021semi}      &     86.81      &       87.70 \\
UGCL(U-Net) \cite{wang2022uncertainty}      &     72.67      &    79.48 \\
\midrule
UCMT(U-Net) (ours)    &     \textbf{82.14}      &     \textbf{83.33}      \\
CMT (ours)    &     87.86      &     88.10      \\ 
UCMT (ours)    &     \textbf{88.22}      &     \textbf{88.46}      \\ \bottomrule[1pt]
\end{tabular}
\caption{Comparison with state-of-the-art methods on ISIC dataset. 5\% $\mathcal{D_L}$ and  10\% $\mathcal{D_L}$ of the labeled data are used for training, respectively. Results are measured by DSC.}
\label{Tab:isic}
\end{table}

\begin{table}[!t]
\centering
\resizebox{\columnwidth}{!}{
\begin{tabular}{lll|ll}
\toprule[1pt]
\multirow{2}{*}{Method} & \multicolumn{2}{c|}{Kvasir-SEG} & \multicolumn{2}{c}{CVC-ClinicDB} \\ \cline{2-5}
& 15\% $\mathcal{D_L}$ & 30\% $\mathcal{D_L}$ & 15\% $\mathcal{D_L}$ & 30\% $\mathcal{D_L}$\\ \midrule
AdvSemSeg      &    56.88       &    76.09     &    68.39       &    75.93          \\
ColAdv      &    76.76       &    80.95     &    82.18       &    \textbf{89.29}          \\
MT     &    87.44       &    88.72    &    84.19       &     84.40         \\
CCT     &     81.14     &      84.67   &    74.20     &      78.46     \\
CPS       &     86.44     &       88.71   &     85.34      &       86.69   \\ \midrule
CMT (ours)    &     88.08     &     88.61     &     85.88      &     86.83      \\ 
UCMT (ours)    &     \textbf{88.68}      &     \textbf{89.06}      &     \textbf{87.30}      &     87.51 \\ \bottomrule[1pt]
\end{tabular}}
\caption{Comparison with state-of-the-art methods on Kvasir-SEG and CVC-ClinicDB datasets in terms of DSC. 15\% $\mathcal{D_L}$ and  30\% $\mathcal{D_L}$ of the labeled data are individually used for training.}
\label{Tab:colon}
\end{table}

\begin{table*}[ht]
\centering
\begin{tabular}{lllll|llll}
\toprule[1pt]
\multirow{2}{*}{Method} & \multicolumn{4}{c|}{10\% $\mathcal{D_L}$} & \multicolumn{4}{c}{20\% $\mathcal{D_L}$} \\ \cline{2-9}
& DSC & Jaccard & 95HD & ASD  & DSC & Jaccard & 95HD & ASD  \\ \midrule
UA-MT \cite{yu2019uncertainty}     &    84.25       &    73.48     &    13.84      &    3.36      &    88.88       &    80.21     &    7.32      &    2.26 \\
SASSNet \cite{li2020shape}     &    87.32       &    77.72     &    9.62       &    2.55       &    89.54       &    81.24     &    8.24       &   2.20      \\
LG-ER-MT \cite{hang2020local}     &   85.54       &    75.12    &    13.29       &     3.77  &   89.62      &    81.31    &    7.16      &    2.06   \\
DUWM \cite{wang2020double}     &    85.91     &      75.75   &     12.67     &      3.31     &    89.65    &    81.35   &     7.04     &      2.03 \\
DTC \cite{luo2021semi}     &    86.57     &      76.55   &     14.47     &      3.74   &    89.42     &      80.98   &     7.32     &      2.10 \\ 
MC-Net \cite{wu2021semi}      &     87.71    &      78.31   &     9.36      &       \textbf{2.18}  &     90.34    &      82.48   &     \textbf{6.00}     &       1.77 \\ 
MT \cite{tarvainen2017mean}      &    86.15   &      76.16   &     11.37      &       3.60  &     89.81    &      81.85   &    6.08     &       1.96 \\ 
CPS \cite{chen2021semi}      &    86.23    &      76.22   &     11.68      &       3.65  &     88.72    &      80.01   &    7.49     &       1.91 \\ \midrule
CMT (ours)    &     87.23     &     77.83     &     \textbf{7.83}      &     2.23  &   89.88     &     81.74     &     6.07      &     1.94     \\ 
UCMT (ours)    &     \textbf{88.13}      &     \textbf{79.18}      &     9.14      &    3.06  &     \textbf{90.41}      &     \textbf{82.54}      &     6.31      &    \textbf{1.70}\\ \bottomrule[1pt]
\end{tabular}
\caption{Comparison with state-of-the-art methods on LA dataset. 10\% $\mathcal{D_L}$ and 20\% $\mathcal{D_L}$ of the labeled data are used for training.}
\label{Tab:la}
\end{table*}

\textbf{Results on Colonoscopy.}
We further conduct a comparative experiment on the polyp segmentation task from colonoscopy images. Table \ref{Tab:colon} reports the quantitative results on both Kvasir-SEG and CVC-ClinicDB datasets. Compared with the adversarial learning-based \cite{hung2018adversarial,wu2021collaborative} and consistency learning-based \cite{tarvainen2017mean,ouali2020semi,chen2021semi} algorithms, the proposed methods achieve the state-of-the-art performance. For example, both CMT and UCMT outperform AdvSemSeg \cite{hung2018adversarial} and ColAdv \cite{wu2021collaborative} by large margins on Kvasir-SEG and CVC-ClinicDB, except that ColAdv shows the better performance of 89.29\% on CVC-ClinicDB under 30\% labeled data. These results demonstrate that our uncertainty-guided collaborative mean-teacher scheme performs better than the adversarial learning and consistency learning schemes commonly used in the compared approaches. 
Notably, CMT and UCMT show better performance on the low-data regime, i.e., 15\% $\mathcal{D_L}$, and the performance between 15\% $\mathcal{D_L}$ and 30\% $\mathcal{D_L}$ labeled data is close.
This phenomenon reflects the capacity of our method to produce high-quality pseudo labels from unlabeled data for semi-supervised learning, even with less labeled data.

\textbf{Results on Cardiac MRI.} 
We further evaluate the proposed method in the 3D medical image segmentation task. Table \ref{Tab:la} shows the comparison results on the 3D left atrium segmentation from cardiac MRI. 
The compared approaches are all based on consistency learning and pseudo labeling, including uncertainty-aware \cite{yu2019uncertainty,wang2020double}, shape-aware \cite{li2020shape}, structure-aware \cite{hang2020local}, dual-task \cite{luo2021semi}, and mutual training \cite{wu2021semi} consistency.
It can be observed that UCMT achieves the best performance under both 10\% and 20\% $\mathcal{D_L}$ in terms of DSC and Jaccard over the state-of-the-art methods. 
For example, compared with UA-MT \cite{yu2019uncertainty} and MC-Net \cite{wu2021semi}, UCMT shows 3.88\% DSC and 0.43\% DSC improvements on the 10\% labeled data. The results demonstrate the superiority of our UCMT for 3D medical image segmentation.

\subsection{Ablation Study}
We conduct an ablation study in terms of network architectures, loss functions and region mix to investigate the effectiveness of each component and analyze the hyperparameters of the proposed method. 
There are three types of network architectures: 1) teacher-student (TS), 2) student-student (SS), and 3) student-teacher-student in the proposed CMT.

\textbf{Effectiveness of Each Component.} 
Table \ref{Tab:ablation} reports the performance improvements over the baseline. It shows a trend that the segmentation performance improves when the components, including the STS (student-teacher-student), $\mathcal{L}_{cps}$, $\mathcal{L}_{mts}$, and UMIX are introduced into the baseline, and again confirms the necessity of   
encouraging model disagreement and enhancing the quality of pseudo labels for semi-supervised segmentation. The semi-supervised segmentation model is boosted for two reasons: 1) $\mathcal{L}_{cps}$, $\mathcal{L}_{mts}$ and the STS architecture that force the model disagreement in CMT for co-training, and 2) UMIX facilitating the model to produce high-confidence pseudo labels.
All the components contribute to UCMT to achieve 88.22\% DSC. These results demonstrate their effectiveness and complementarity for semi-supervised medical image segmentation. 
On the other hand, the two groups of comparisons between "TS (teacher-student) + $\mathcal{L}_{mts}$" (i.e., MT) vs. STS + $\mathcal{L}_{mts}$ (i.e., CMTv1), and between "SS (student-student) + $\mathcal{L}_{cps}$" (i.e., CPS) vs. "STS + $\mathcal{L}_{cps}$" (i.e., CMTv2) show that the STS-based approaches yield the improvements of  0.17\%  and  0.64\%, indicating the effectiveness of the STS component. 
However, the performance gaps are not significant because the STS architecture increases the co-training disagreement but decreases the confidences of pseudo labels. 
It can be easily found that 
 the results are improved to 87.86\% by "STS + $\mathcal{L}_{cps}$ + $\mathcal{L}_{mts}$" (i.e., CMTv3) and 
 the relative improvements of 1.55\% and 1.41\% DSC have been obtained by "STS + $\mathcal{L}_{cps}$ + $\mathcal{L}_{mts}$ + UMIX" (i.e., UCMT) compared with MT and CPS.
 The results demonstrate our hypothesis that maintaining co-training with high-confidence pseudo labels can improve the performance of semi-supervised learning.

 \begin{table}[!t]
\centering
\begin{tabular}{c|ccc|cc|c|c}
\toprule[1pt]
Method & TS & SS & STS & $\mathcal{L}_{cps}$ & $\mathcal{L}_{mts}$ &  U  &  DSC \\
\midrule
Baseline	&    &   &   &    &   &   & 83.31 \\
MT	&  $\surd$  &   &   &    &  $\surd$ &   & 86.67 \\
CPS	&    & $\surd$  &   &  $\surd$ &    &   & 86.81 \\
CMTv1 &   &   &   $\surd$ &   & $\surd$  &   & 86.84 \\
CMTv2 &  &   &   $\surd$  & $\surd$  &   &   & 87.48 \\
CMTv3 &   &   &  $\surd$  &  $\surd$ & $\surd$  &   & 87.86 \\
UCMT  &   &  &  $\surd$  &  $\surd$ & $\surd$  & $\surd$ & \textbf{88.22} \\
\bottomrule[1pt]
\end{tabular}
\caption{Ablation study of the different component combinations with only 5\% labeled data on ISIC dataset. TS: teacher-student; SS: student-student; STS: student-teacher-student; $\mathcal{L}_{cps}$: cross pseudo supervision; $\mathcal{L}_{mts}$: mean-teacher supervision; U: UMIX.}
\label{Tab:ablation}
\end{table}

\begin{figure}[!t]
\centering
\includegraphics[width=3.4 in]{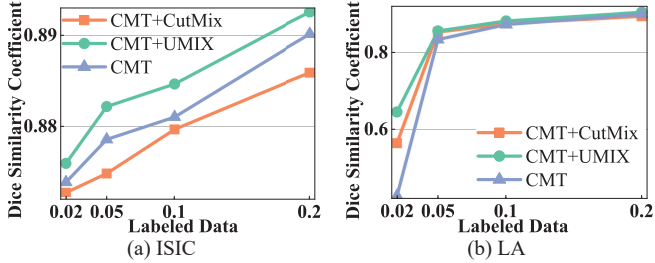}
	\caption{Comparison of UMIX and CutMix on ISIC (a) and LA (b) dataset under 2\%, 5\%, 10\%, and 20\% $\mathcal{D_L}$.}
	\label{FIG:Line_ISIC_LA}
\end{figure}

\textbf{Comparison of Different Data Augmentation.}
We further compare the proposed UMIX, component of our UCMT, with CutMix \cite{yun2019cutmix} on ISIC and LA datasets with different labeled data to investigate their effects in semi-supervised segmentation. As illustrates in Figure \ref{FIG:Line_ISIC_LA}, UMIX outperforms CutMix, especial in the low-data regime, i.e., 2\% labeled data.
The reason for this phenomenon is that CutMix performs random region mix that inevitably introduces noise into the new samples, which reduces the quality of the pseudo labels, while UMIX processes the image regions according to the uncertainty of the model, which facilitates the model to generate more confident pseudo labels.

\textbf{Parameter Sensitivity Analysis.}
UMIX has two hyperparamters, i.e., the top $k$ regions for mix and the size of the regions (patches) defined as the ratio of $1/r$ to the image size.
We study the influence of these factors to UCMT on ISIC dataset with 5\% $\mathcal{D_L}$.
It can be observed in Table \ref{Tab:topk} that reducing the patch size leads to a slight increase in performance.
Moreover, varying the number of $k$ does not bring us any improvement, 
suggesting that we can choose any value of $K$ to eliminate outliers, thus bringing high-confidence pseudo labels for semi-supervised learning, which indicates the robustness of UMIX.

\begin{table}[t]
\centering
\begin{tabular}{l|ccccc}
\toprule[1pt]
\diagbox{$1/r$}{$k$}  & 1 & 2 & 3 & 4 & 5\\ \midrule
1/16     &87.95 &\textbf{88.22}  &88.12  &87.96 &88.08 \\
1/4      &87.65 &88.15  &87.80 &87.54 &87.90\\ 
1/8    &87.86 &87.92  &88.03 &88.01 &87.87\\  
\bottomrule[1pt]
\end{tabular}
\caption{Investigation on how the top $k$ and patch size affect the capacity of UMIX on ISIC dataset with 5\% labeled data.}
\label{Tab:topk}
\end{table}

\subsection{Qualitative Results}
\begin{figure}[!t]
\centering
\includegraphics[width=3.0 in]{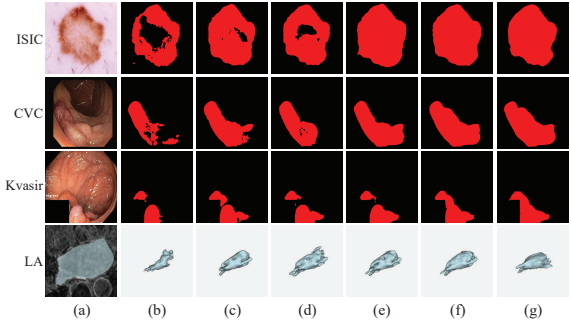}
	\caption{Qualitative examples on the four datasets. (a) images, (b) supervised baseline, (c) MT, (d) CPS, (e) CMT, (f) UCMT, and (g) ground truth. }
	\label{FIG:qualitative}
\end{figure}

\begin{figure}[!t]
\centering
\includegraphics[width=3.0 in]{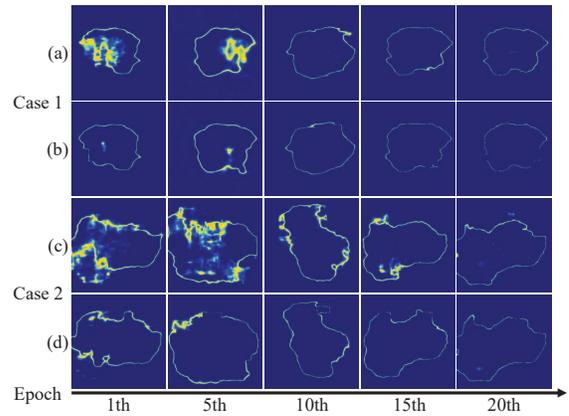}
	\caption{Illustration of the uncertainty maps. (a) and (b) are the uncertainty maps of the original images and the UMIX images for the first example, while (c) and (d) are for the second example.}
	\label{FIG:umaps}
\end{figure}

Figure \ref{FIG:qualitative} visualizes some example results of polyp segmentation, skin lesion segmentation, and left atrial segmentation. As shown in Figure \ref{FIG:qualitative} (a), the supervised baseline insufficiently segments some lesion regions, mainly due to the limited number of labeled data. 
Moreover, MT [Figure \ref{FIG:qualitative} (b)] and CPS [Figure \ref{FIG:qualitative} (c)] typically under-segment certain objects, which can be attributed to the limited generalization capability. On the contrary, our CMT [Figure \ref{FIG:qualitative} (e)] corrects these errors and produces smoother segment boundaries by  gaining more effective supervision from unlabeled data. Besides, our complete method UCMT [Figure \ref{FIG:qualitative} (f)] further generates more accurate results by recovering finer segmentation details through more efficient training. These examples qualitatively verify the robustness of the proposed UCMT.
In addition, to clearly give an insight into the procedure of the pseudo label generation and utilization in the co-training SSL method, we illustrate the uncertainty maps for two samples during the training in Figure \ref{FIG:umaps}. As shown, UCMT generates the uncertainty maps with high uncertainty [Figure \ref{FIG:umaps} (a)/(c)] in the early training stage whereas our model produces relative higher confidence maps [Figure \ref{FIG:umaps} (b)/(d)] from the UMIX images. During training, UCMT gradually improves the confidence for the input images. These results prove that UMIX can facilitate SSL models to generate high-confidence pseudo labels during training, guaranteeing that UCMT is able to maintain co-training in a more proper way.

\section{Conclusion}
We present an uncertainty-guided collaborative mean-teacher for semi-supervised medical image segmentation. Our main ideas lies in maintaining co-training with high-confidence pseudo labels to improve the capability of the SSL models to explore information from unlabeled data. Extensive experiments on four public datasets demonstrate the effectiveness of this idea and show that the proposed UCMT can achieve state-of-the-art performance. In the future, we will investigate more deeply the underlying mechanisms of co-training for more effective semi-supervised image segmentation.

\appendix

\section*{Ethical Statement}

There are no ethical issues.

\section*{Acknowledgments}
This work was supported by the National Natural Science Foundation of China under Grant 62076059 and the Natural Science Foundation of Liaoning Province under Grant 2021-MS-105.

\bibliographystyle{named}
\bibliography{ijcai23}

\end{document}